\documentclass[letterpaper, 10 pt, conference]{ieeeconf}  

\IEEEoverridecommandlockouts                              

\usepackage{amsmath}
\usepackage{amsfonts}
\usepackage{algorithm}
\usepackage{algpseudocode}
\usepackage{graphicx}
\usepackage{subcaption}
\usepackage{lipsum}
\usepackage{caption}
\usepackage{cite}
\graphicspath{{images/}}

\title{\LARGE \bf
Deep Learning with Experience Ranking Convolutional Neural Network for Robot Manipulator}
\vspace{-10pt}
\author{Hai Nguyen, Hung Manh La, and Matthew Deans  
\thanks{This material is based upon work supported by the National Aeronautics and Space Administration (NASA) Grant No. NNX15AI02H issued through the NVSGC Research Infrastructure and the NVSGC Faculty Award - CD.}
\thanks{Hai Nguyen and Dr. Hung La are with the Advanced Robotics and Automation
(ARA) Laboratory, Department of Computer Science and Engineering, University of Nevada, Reno, NV 89557, USA. Dr. Matthew Deans is with NASA Ames Research Center, Moffett Field, CA 94035. Corresponding author: Hung La, email: {\tt\small hla@unr.edu}}
}

\begin{document}

\maketitle
\thispagestyle{empty}
\pagestyle{empty}

\vspace{-10pt}
\begin{abstract}
Supervised learning, more specifically Convolutional Neural Networks (CNN), has surpassed human ability in some visual recognition tasks such as detection of traffic signs, faces and handwritten numbers. On the other hand, even state-of-the-art reinforcement learning (RL) methods have difficulties in environments with sparse and binary rewards. They requires manually shaping reward functions, which might be challenging to come up with. These tasks, however, are trivial to human. One of the reasons that human are better learners in these tasks is that we are embedded with much prior knowledge of the world. These knowledge might be either embedded in our genes or learned from imitation - a type of supervised learning. For that reason, the best way to narrow the gap between machine and human learning ability should be to mimic how we learn so well in various tasks by a combination of RL and supervised learning. Our method, which integrates Deep Deterministic Policy Gradients and Hindsight Experience Replay (RL method specifically dealing with sparse rewards) with an experience ranking CNN, provides a significant speedup over the learning curve on simulated robotics tasks. Experience ranking  allows high-reward transitions to be replayed more frequently, and therefore help learn more efficiently. Our proposed approach can also speed up learning in any other tasks that provide additional information for experience ranking.
\end{abstract}

\vspace{-2pt}
\section{Introduction}

Reinforcement Learning (RL) methods have been applied to a variety of real-world robotics problems ranging from locomotion \cite{DRL1} to manipulation \cite{DRL4}, \cite{DRL6}. However, one of the main challenges that limits RL's full potential for robotics applications is how to learn efficiently from sparse and binary rewards, which is very common in many robotics tasks. It is challenging as in most learning attempts, by exploration, the agents will mostly end up learning nothing from zero rewards, and they are clueless about what to improve next time. One of the best methods that we have had so far to tackle the problem is using Hindsight Experience Replay (HER) \cite{andrychowicz2017hindsight} in combination with an off-policy RL methods e.g., Deep Deterministic Policy Gradient (DDPG) \cite{DDPG}, Normalized Advantage Functions (NAF) \cite{NAF}. By modifying the original goal, HER allows the agent to even learn from unsuccessful transitions.  

One problem with HER is that it used a hard-coded strategy (\textit{final} or \textit{future} strategy) \cite{HERReport} for randomly choosing additional goals and episodes to replay. This lack of an evaluation scheme makes it possible that the chosen transitions are poor experiences, which might slow the learning process. To solve this problem and respond to OpenAI's request for research for an automatic hindsight goal generation for HER mentioned in  \cite{HERReport}, we propose a novel idea of using an external CNN network to evaluate the training experience. Based on their ranking scores, we will decide which episodes should be restored and replayed for training. 

We verified that the proposed ranking scheme can improve significantly the learning curve for a mobile robot manipulator to learn four different tasks in robotics environment of Gym \cite{Gym}. Our proposed approach will also be useful for any other robotics learning problems that can provide additional information for experience ranking such as images.

\vspace{-5pt}
\section{Related Work}
\label{Content:Related Work}
In biological systems, experience replay has played an important role for learning behaviors. Various studies show that there is frequent experience replay in the hippocampus of rodents of both awake and sleeping animals. Research also shows that disrupting experience replay might lead to impairing spatial memory \cite{Girardeau2009} \cite{Ego-Stengel}, and some show that experience replay can contribute to better memory consolidation and retrieval \cite{Carr2011}, both are instrumental in boosting learning ability.

In machine learning, the idea of using \textit{Experience Replay} was first introduced in \cite{Lin1992SelfImprovingRA} when it is used to speed up learning by repeating past experiences. Experience replay helps break the temporal correlation by mixing old and new experiences, and frequent experiences therefore might be replayed more often. Since then, experience replay has been widely adapted, and it becomes the norm in many success in RL research \cite{mnih2013playing} \cite{mnih2015humanlevel}.

However, the current way to use experience replay is to uniformly sampled from first-in-first-out (FIFO) buffers of experiences \cite{DDPG} \cite{mnih2015humanlevel}. By doing this way, the importance of each transition is neglected, and we might replay experience that is not very helpful for the learning at all. When faced with the design choice of which experiences to restore, neuroscience research show that surprising \cite{CHENG2008303} and rewarding experiences are more preferred \cite{ATHERTON2015560}. 

In RL, attempts to prioritized experiences have been made in \cite{PER} where temporal difference (TD) error is used as a way to measure the priority. Our approach is reward-favored and more direct. We will look at the state of the training episode and use a pre-trained neural network to rank that episode and decide whether the associated experience is worth being stored for future learning.

\section{Background}
\label{Content:Background}
\boldmath

\subsection{Reinforcement Learning Formalism}
Considering the standard reinforcement learning setting when an agent interacts with an environment. The agent continually makes decisions, and the environment responds to the chosen actions and bring agents to new states. At each step, the agent receives the current environment's state $S_t \in \mathcal{S}$ and bases on the current policy $\pi$ to select an action $a_t \in \mathcal{A}$. The next time step, the agent receives a numerical reward $R_{t+1} \in \mathbb{R}$ with a reward function $r: \mathcal{S} \times \mathcal{A} \rightarrow \mathbb{R}$ and itself being in another state $S_{t+1}$. The goal of the agent is to maximize the reward over time through learning a policy, which can help to choose actions that lead to maximizing the accumulated reward. The dynamics of the environment is defined by the state-transition probability $p :\mathcal{S}\times\mathcal{S}\times\mathcal{A}\rightarrow [0, 1]$.

A deterministic policy will map states to actions: $\pi:\mathcal{S}\rightarrow\mathcal{A}$. Every episode starts with sampling an initial state $s_{0}$, and at every timestep $t$ the agent selects an action based on the current policy and its current state: $a_t = \pi(s_t)$. Then it receives the reward $r_t=r(s_t,a_t)$, and the environment's new state is sampled from the distribution $p(.|s_t,a_t)$. 

A discounted sum of future rewards is $R_{t} = \sum_{i=t}^{\infty} \gamma^{i-t}r_i$ with a discount rate $\gamma \in [0,1]$, and the agent's goal is to maximize its expected return starting from the beginning $\mathbb{E}_{s_0}[R_0|s_0]$. 

The action-value function describing the value of taking action $a$ in state $s$ under the policy $\pi$ is defined as $Q^\pi(s,a)=\mathbb{E}[R_t|s_t=s,a_t=a]$. An optimal policy $\pi^*$ i.e. any policy $\pi^*$ s.t. $Q^{\pi^*}(s,a) \geq Q^\pi(s,a)$ for every $s \in \mathcal{S}, a \in \mathcal{A}$, and any policy $\pi^*$ share the same \textit{optimal Q-function} denoted as $Q^*$. It satisfies the \textit{Bellman} equation:
\begin{equation*}
Q^*(s,a)=\mathbb{E}\Big[R(s,a)+\gamma \textit{max} Q^*(s',a')\Big].
\end{equation*}

\subsection{Deep Q-Networks (DQN)}
DQN \cite{mnih2015humanlevel} is a model-free RL algorithm for discrete action spaces. In DQN, neural networks are used to approximate the action-value function $Q(s,a)$. Two important features of DQN are \textit{experience replay} and \textit{target network}. 

\textit{Experience replay} is used to store previously transitions, which are defined as tuples $(s_t, a_t, r_t, s_{t+1})$ where $s_t$ is the current state, $a_t$ is the action, $r_t$ is the reward, and $s_{t+1}$ is the next state. Tuples will then be added to a \textit{replay buffer} from which batch of transitions will be sampled for training. The usage of experience replay provides uncorrelated samples for training and also improve data efficiency \cite{Wang}.

The network is then trained using mini-batch gradient descent on the loss $\mathcal{L} = \mathbb{E}(Q(s_t,a_t)-y_t)^2$, where $y_t=r_t+\gamma \textit{max}_{a' \in A} Q(s_{t+1},a')$, with the tuple $(s_t,a_t,r_t,s_{t+1})$ sampled from the replay buffer.

In order to make this optimization procedure stabler, the target $y_t$ is usually computed using a separate \textit{target network}, which changes at a slower pace than the main network. After that, the weights of the \textit{target network} will be periodically updated to the main network to increase the stability.

\subsection{Deep Deterministic Policy Gradient - DDPG}

Although DQN has achieved good performance in higher dimensional problems such as Atari games \cite{mnih2013playing}, it only works with discrete action spaces. DDPG \cite{DDPG} is developed based on DQN, and it is a model-free off-policy RL algorithm for continuous action spaces. 

DDPG belongs to the actor-critic structure where there are two neural networks: an \textit{actor} $\pi: \mathcal{S} \rightarrow \mathcal{A}$ and a \textit{critic} $Q: \mathcal{S} \times \mathcal{A} \rightarrow \mathbb{R}$. The critic's job is to evaluate the current policy by estimating the Q-value from the current state and the action outputted from the actor.

Firstly, action is generated by the target policy with added noise for action exploration, e.g., $a_t = \pi(s) + \mathcal{N}_t$. The critic is trained in a similar way as the Q-function in DQN with the target $y_t$ being computed using actions produced by the actor, e.g., $y_t=r_t+\gamma Q(s_{t+1}, \pi(s_{t+1}))$. The actor is trained using mini-batch gradient descent applied on the loss $\mathcal{L} = -\mathbb{E}_sQ(s,\pi(s))$.

\subsection{Hindsight Experience Replay - HER}
The idea behind HER is to mimic human ability to learn from failures. HER allows learning from all episodes , even if in those episodes the agent did not achieve the original goal. Instead, HER considers the final state that the agent finally reached to be a modified goal. While in the standard experience replay, only the transition $(s_t||g, a_t, r_t, s_{t+1}||g)$ with the original goal $g$ is stored, HER also keeps the transition $(s_t||g', a_t, r', s_{t+1}||g')$ with modified goal $g'$. By adding additional goals, the RL algorithm can still have a learning outcome because it can learn to achieve a goal even that goal is not the original target.

In \cite{andrychowicz2017hindsight}, they experimented with different strategies for choosing goals to use with HER: \textit{final} strategy - goals corresponding to the final state of the environment and \textit{future}: \textit{k} random states which come from the same episode as the transition being replayed and were observed \textit{after} it. However, both strategies neglect the importance of each episode, and there is actually no mechanism to measure.

\begin{figure*}[htb]
  \centering
  \begin{minipage}{2\columnwidth}
  \begin{algorithm}[H]
  \caption{DDPG + HER modified with ER-CNN}  
  \label{Alg:DDPG+HER+CNN}
  \begin{algorithmic}[1]
\State Initialize DDPG
\State Initialize replay buffer $R$
\For{episode = 1, $M$}
\State Sample a goal $g$ and an initial state $s_0$ 
    \For{$t$ = 0, $T$-1}
    \State Sample an action $a_t$ using DDPG behavioral policy
    \State Execute the action $a_t$ and observe a new state $s_{t+1}$
    \EndFor
    \For{$t$ = 0, $T$ - 1}
    \State $r_t := r(s_t, a_t, g)$
    \State Store the transition $(s_t||g, a_t, r_t, s_{t+1})$ in $R$
    \State Sample a set of additional goals for replay $G := \mathbb{S}$ \textbf{(current episode)}
    	\For{$g' \in G$}
    	\State Retrieve the rank of the transition from ER-CNN \Comment ER-CNN to rank the experience
    	\If{rank $\geq$ threshold} \Comment{Choose highest ranked transitions to store}
    		\State $r' := r(s_t, a_t, g')$
    		\State Store the transition $(s_t||g', a_t, r', s_{t+1})$ in $R$ \Comment{HER}
    	\Else
    	\State Next loop \Comment{Pass the current transition}
    	\EndIf
    	\EndFor
    \EndFor
    \For{$t$ = 1, $N$}
    \State Sample a minibatch $B$ from the replay buffer $R$
    \State Perform one step of optimization 
    \EndFor
\EndFor

  \end{algorithmic}
  \end{algorithm}
  \end{minipage}
\end{figure*}

\section{DDPG + HER and Experience Ranking CNN}
\label{Content:Algorithm}
\subsection{Algorithm}
\begin{figure}[htb]
    \begin{subfigure}[b]{1\columnwidth}
        \includegraphics[width=1\columnwidth]{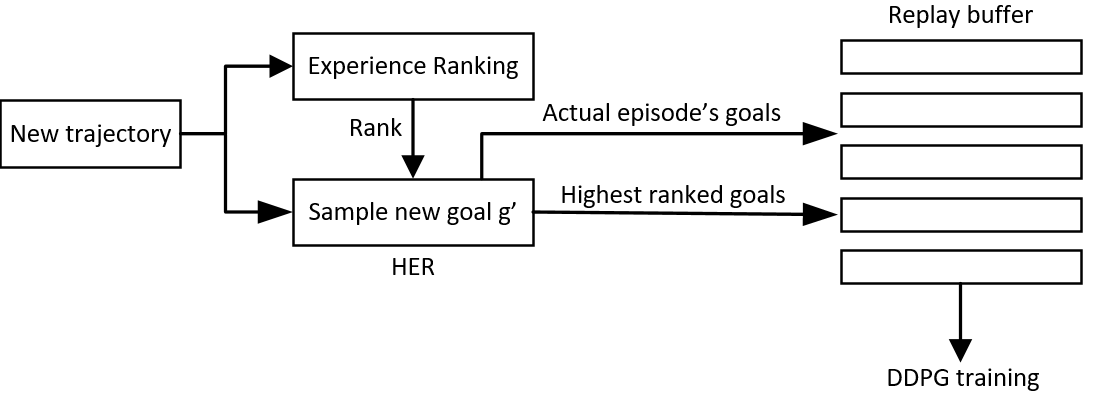}
         \end{subfigure}
         \caption{An overview of the proposed approach for DDPG+HER with Experience Ranking.}
         \label{fig:systemstructure}
         \vspace{-10pt}
\end{figure}

The idea is to improve the sampling of additional goals of HER by introducing new information - ranks of new trajectories as shown in Figure \ref{fig:systemstructure}. Additional goals selected by HER will then contain only episodes that have high rewards. We do not account for the loss of diversity here because HER itself allows a number of percentage of normal episodes to pass through \cite{andrychowicz2017hindsight} by specifying a hyper-parameter \textit{k} controlling the ratio between HER replays and regular replays (e.g. \textit{k} = 4 - 4 times). The output of HER selection algorithm will include both ranked experiences and normal ones.

The modification to DDPG + HER is described in Algorithm \ref{Alg:DDPG+HER+CNN}. In the algorithm, we change HER's random sampling regardless of importance of transitions by using experience ranking to only allow episodes in which the agent achieve high reward to be stored. If the rank of the episode is low, the algorithm jumps to the start of the next \textit{for} loop to sample new experiences.

\subsection{Convolutional Neural Network}
In this section, we briefly describe common knowledge about CNN before continuing with the specific CNN structure that we used to rank the training experiences.

CNN uses various underlying layers consisting of convolution layers, pooling layers and fully connected layers to predict any sort of recommendation, objects in image or video, or some natural language processing for voice recognition \cite{le2017autonomous} \cite{Gibb1}.

\begin{figure}[hbt]
	\centering
    \includegraphics[width=0.5\textwidth]{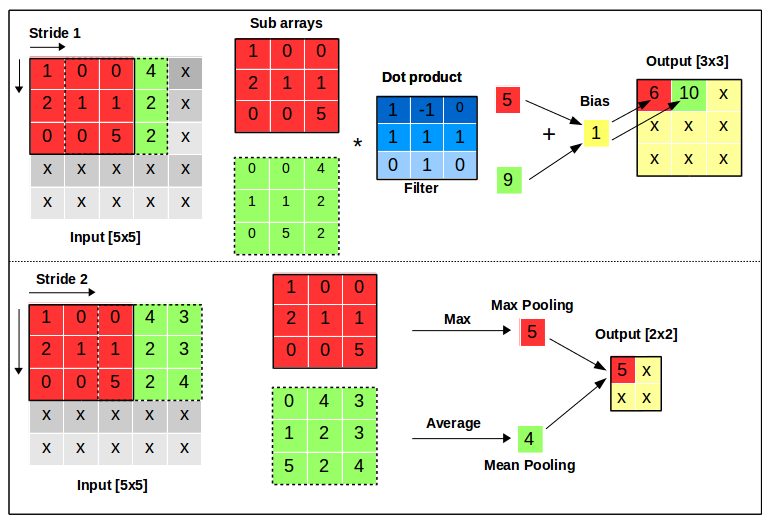}
    \caption{Illustration of convolution and pooling layer.}
    \label{fig:CNN}
    \vspace{-10pt}
\end{figure}

Convolution layers perform the convolution operation between an input image and a set of filters (kernels) whose weights are updated during training \cite{Gibb_IJFR2018}. First, a convolution layer will perform dot product between a sub-array of input (receptive fields) and a filter, whose size is often smaller than the input array. After that, the results will be summed, and a bias term will be added. The way convolution layers work is demonstrated in Figure \ref{fig:CNN}.

Pooling layers often follows preceding convolution operations, and they reduce the spatial size of the input array in so-called down-sampling process. They do not have any trainable parameters. There are two options for pooling layers: max pooling takes the maximum value from the input array, and mean pooling chooses the average values as shown in Figure \ref{fig:CNN}. \par

Fully connected layers have full connection to all activation units of the previous layer and combine learned features from previous lower layers to make a decision based on a high-level reasoning.

\begin{figure} [hbt]
    \begin{subfigure}[b]{1\columnwidth}
        \includegraphics[height=5cm]{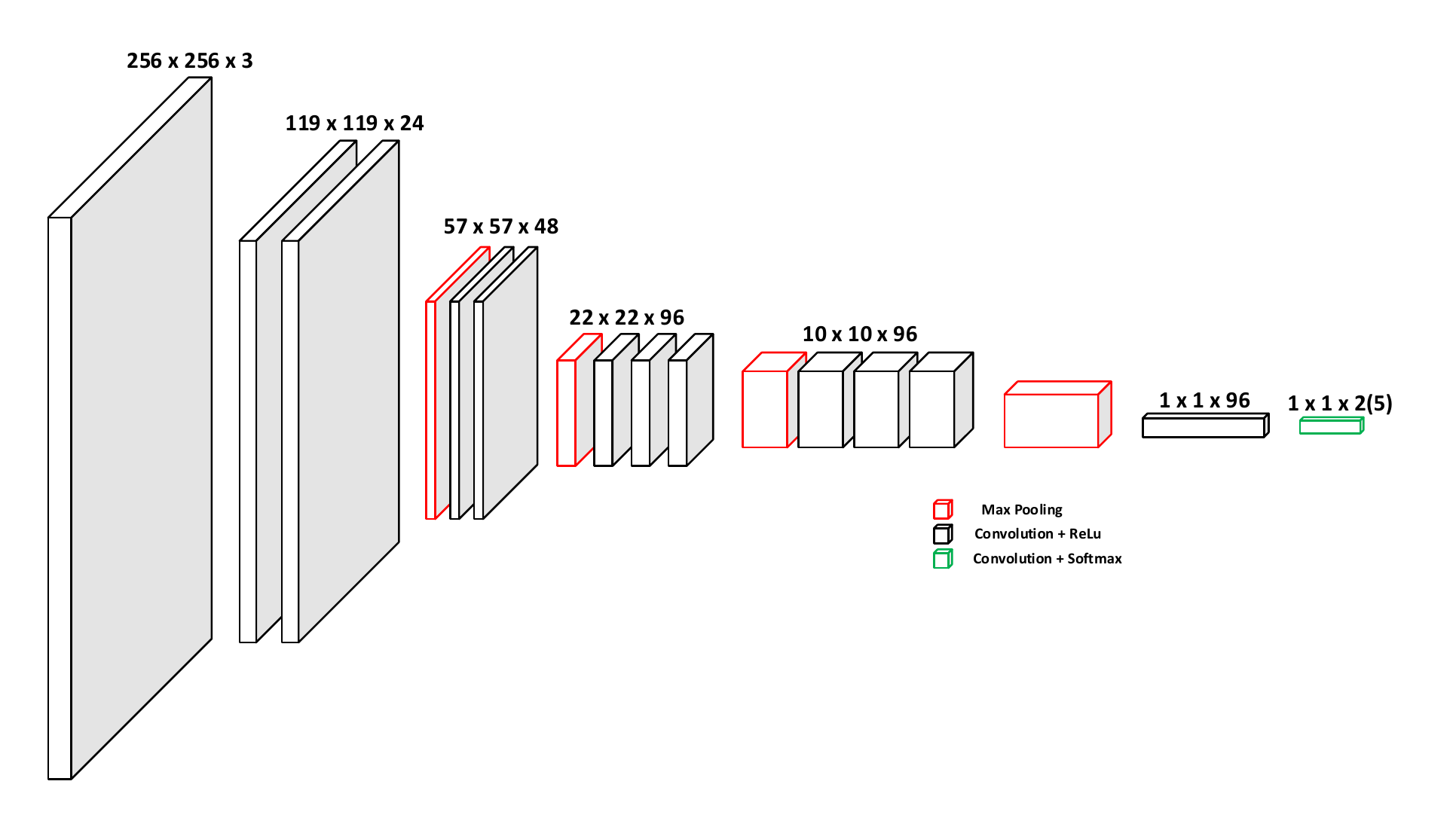}
         \end{subfigure}
         \caption{CNN base line structure.}
         \label{fig:cnnbase}
         \vspace{-10pt}
\end{figure}

In our experiment of a mobile robot manipulator learning to open a door, we used two CNNs: an experience ranking CNN (ER-CNN) for assisting HER by performing evaluation on door-opening training experiences and another for detecting the door handle. In both networks, we use the same baseline structure as shown in Figure \ref{fig:cnnbase}, the only difference between them is the last Convolution layer whose number of outputs will depend on the number of classes. Parameters for all layers used for the two CNNs are shown in Table \ref{Table:LayerParams}.

\begin{table}[hbt]
\begin{center}
 \begin{tabular}{|c | c | c | c|} 
 \hline
 \textit{Layer Type} & \textit{Filter Size} & \textit{Number} & \textit{Stride}\\ [0.5ex] 
 \hline
 Convolution + ReLU & 20x20 & 24 & 2	 \\ 
 \hline
 Max Pooling & 7x7 & N/A & N/A \\
 \hline
 Convolution + ReLU  & 15x15 & 48 & 2\\
 \hline
 Max Pooling & 4x4 & N/A & N/A\\
 \hline
 Convolution + ReLU & 10x10 & 96 & 2\\
 \hline 
 Convolution + Softmax & 1x1 & 2(5 for ER-CNN) & 2\\
 \hline 
\end{tabular}
\end{center}
\caption{Parameters of 2 CNN networks.} 
\label{Table:LayerParams}
\end{table}

\subsubsection{CNN for Experience Ranking}

The purpose of the ER-CNN is to rank the robot's experience during episodes of training. Only episodes with the score being higher than a threshold will be fed to the experience replay buffer. We rank each episode by directly looking at the images showing the states of the door and the manipulator at the end of that training episode. More specifically, if the door's hinge is open more than a 0.05 rad at the end of that episode, the experience is in Group 0 (highest score) and if not, its group (1-4) will be decided based on the distance from the robot's end-effector to the door handle.
\begin{figure*}[htb]
    \centering
    \begin{subfigure}[b]{0.2\textwidth}
        \centering
        \includegraphics[height=2.7cm]{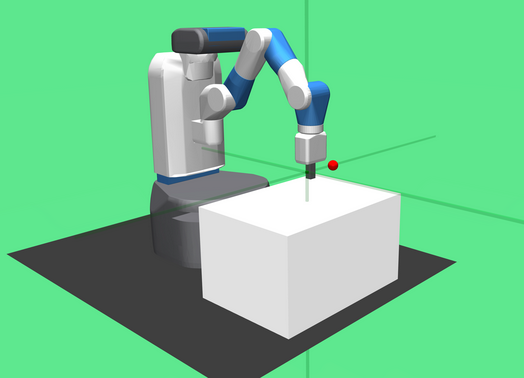}
    \end{subfigure}
    \hspace{\fill}
    \begin{subfigure}[b]{0.2\textwidth}  
        \centering 
        \includegraphics[height=2.7cm]{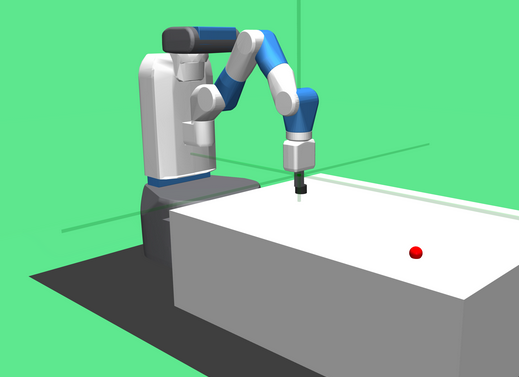}
    \end{subfigure}
    \hspace{\fill}
    \begin{subfigure}[b]{0.2\textwidth}   
        \centering 
        \includegraphics[height=2.7cm]{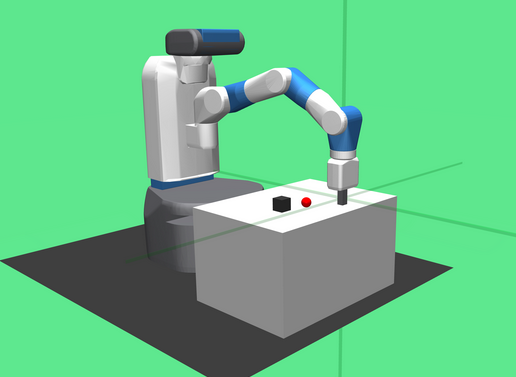} 
    \end{subfigure}
    \hspace{\fill}
    \begin{subfigure}[b]{0.2\textwidth}   
        \centering 
        \includegraphics[height=2.7cm]{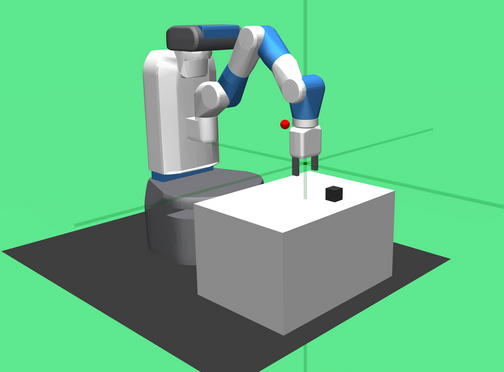}
    \end{subfigure}
	\caption{Fetch Environment. From Left to Right: (1) \textit{FetchReach-v1} Fetch has to move its end-effector to the desired goal position; (2) \textit{FetchSlide-v1} Fetch has to hit a puck across a long table such that it slides and comes to rest on the desired goal; (3) \textit{FetchPush-v1} Fetch has to move a box by pushing it until it reaches a desired goal position; and (4) \textit{FetchPick\&Place-v1} Fetch has to pick up a box from a table using its gripper and move it to a desired goal above the table.}
    \label{fig:FetchEnvApp}
    \vspace{-10pt}
\end{figure*}

This grouping process will be autonomously performed on images captured at the end of training episodes to generate labeled data for training the ER-CNN. The trained ER-CNN will then be able to predict the rank of any episode from images taken at the end of that episode. In this scenario, the trained ER-CNN's role is similar to a human expert supervising the whole training process and giving score for each episode.
\begin{figure}[htb]
    \begin{subfigure}[b]{0.3\columnwidth}
        \centering
        \includegraphics[height=2cm, width = 2.5cm]{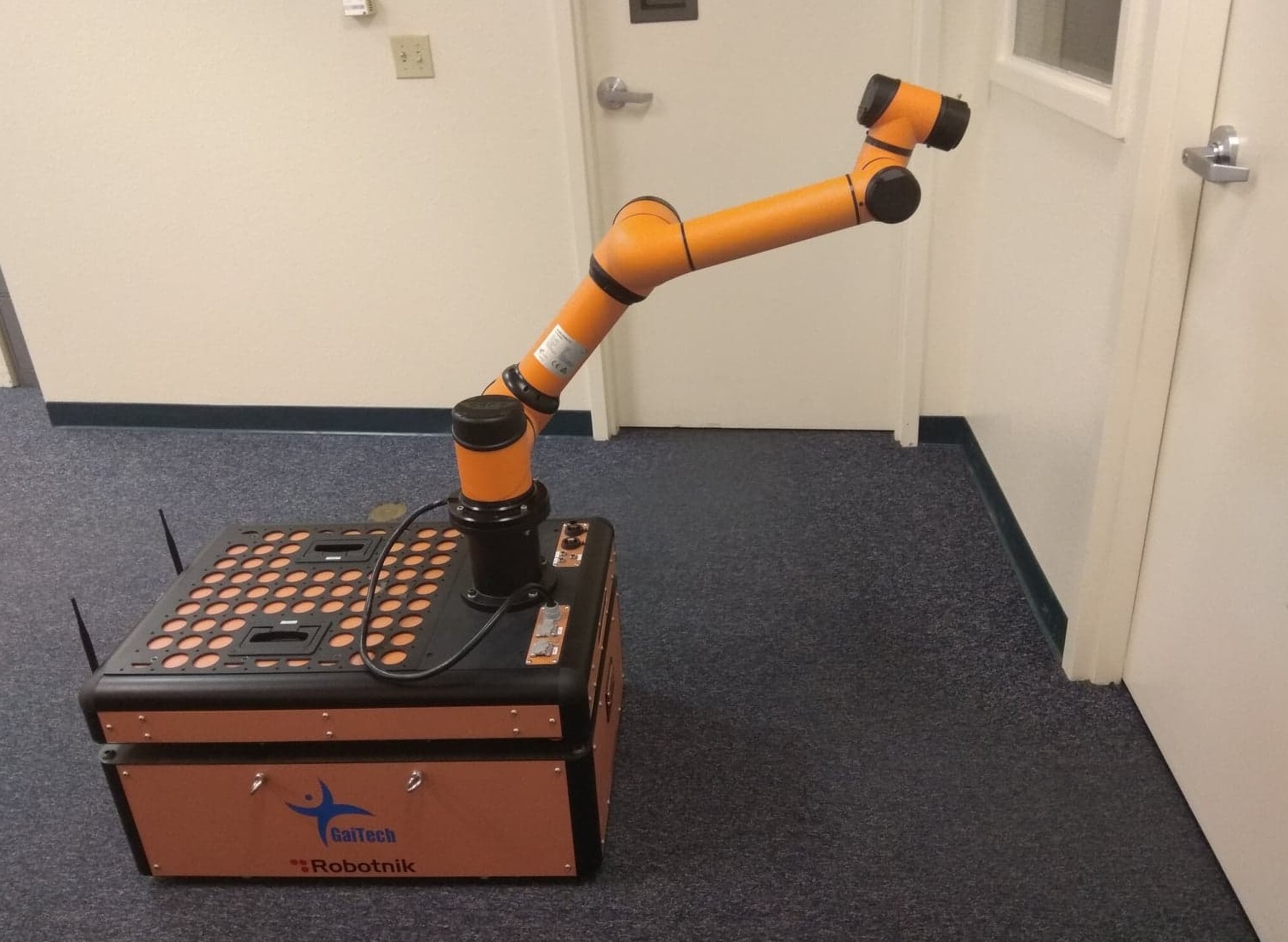}
    \end{subfigure}
    \hspace{\fill}
    \begin{subfigure}[b]{0.3\columnwidth}  
        \centering 
        \includegraphics[height=2cm, width = 2.5cm]{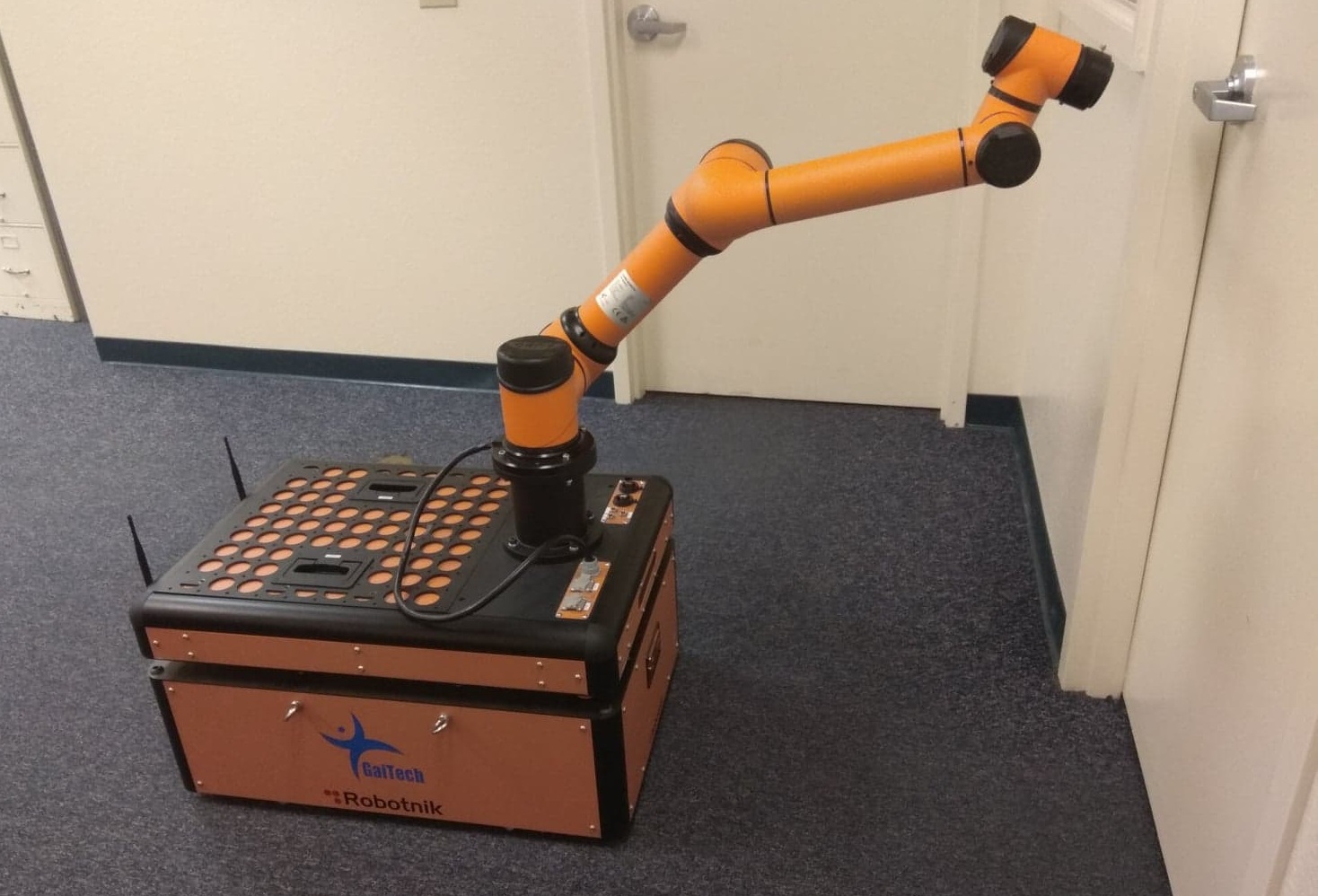}
    \end{subfigure}
    \hspace{\fill}
    \begin{subfigure}[b]{0.3\columnwidth}   
        \centering 
        \includegraphics[height=2cm, width = 2.5cm]{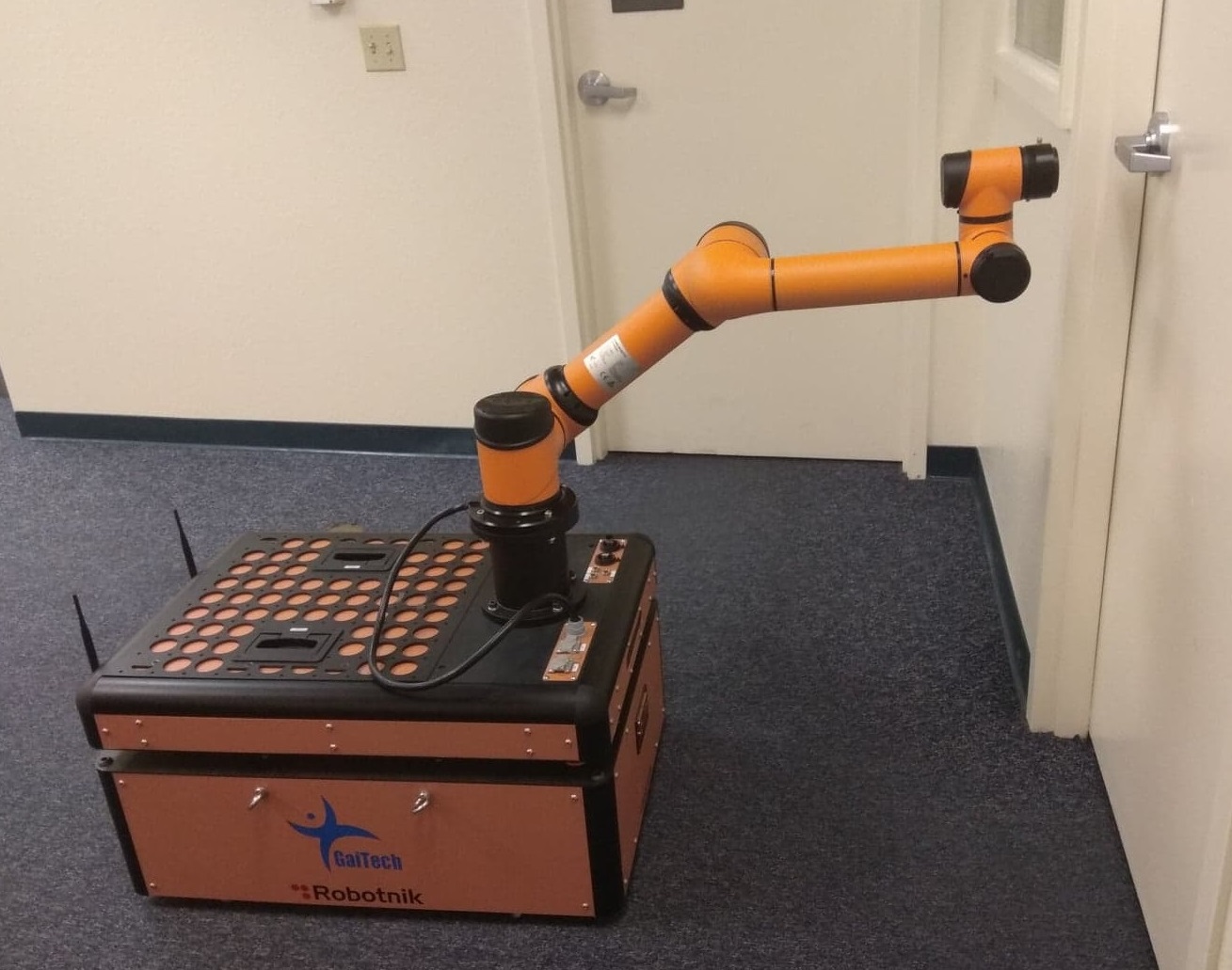} 
    \end{subfigure}
	\caption{Aubo-i5 robot working to open a door.}
    \label{fig:aubo}
\end{figure}

\subsubsection{CNN for Door Handle Detection}

The second CNN will try to detect door handle from images, and the manipulator will push open the door when the detection is positive. The final layer of this CNN will output only two outcomes: detecting the door or not. To further train the network for a better accuracy, we also take the advantage of the image stream coming from the ER-CNN. We perform online training on the pre-trained CNN by applying stochastic gradient descent with the learning rate being slowly decreased for better convergence near the minimum.
\section{Experiment And Results}
\label{Content:Experiment}

\begin{figure*}[htb]
    \centering
    \begin{subfigure}[b]{0.2\textwidth}
        \centering
        \includegraphics[height=4cm]{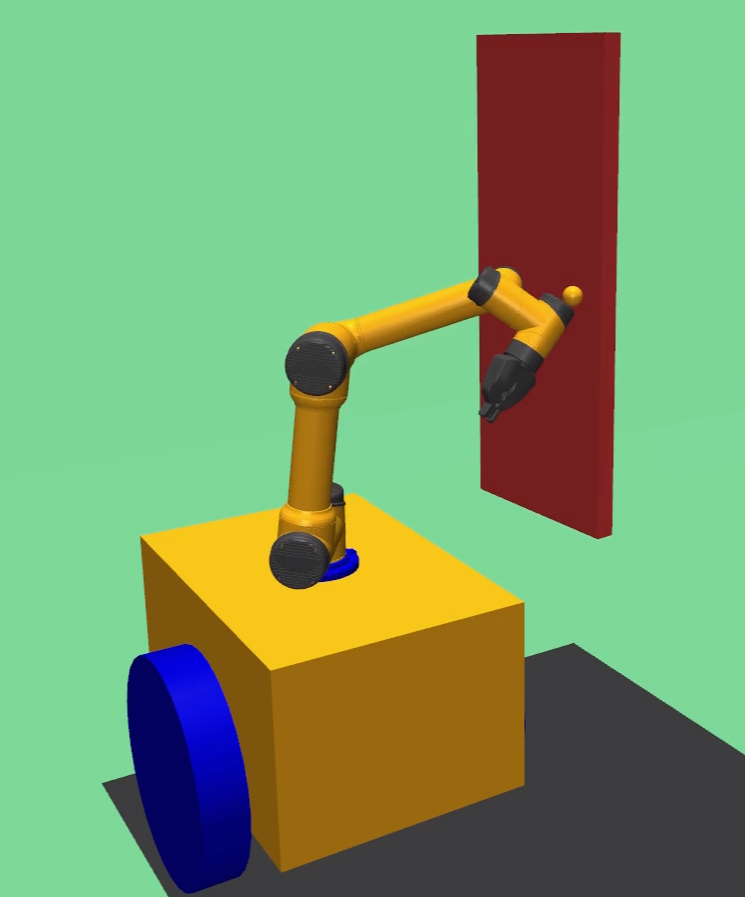}
    \end{subfigure}
    \hspace{\fill}
    \begin{subfigure}[b]{0.2\textwidth}  
        \centering 
        \includegraphics[height=4cm]{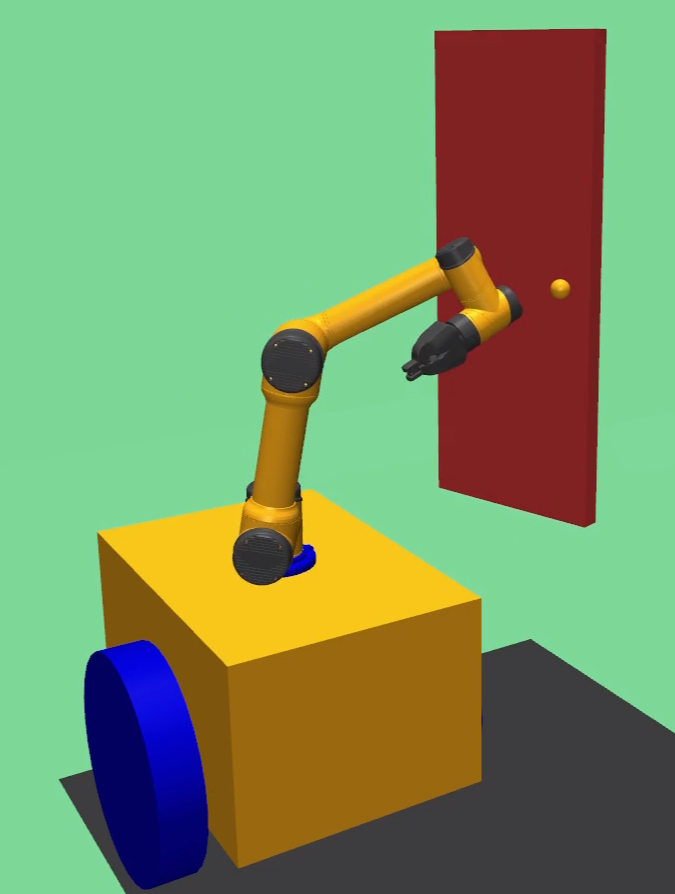}
    \end{subfigure}
    \hspace{\fill}
    \begin{subfigure}[b]{0.2\textwidth}   
        \centering 
        \includegraphics[height=4cm]{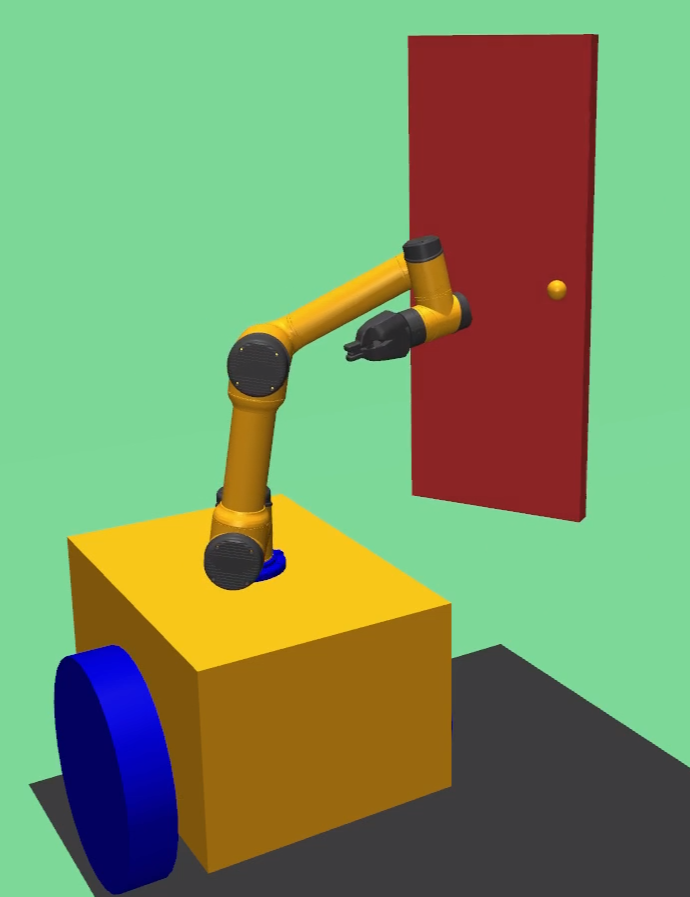} 
    \end{subfigure}
    \hspace{\fill}
    \begin{subfigure}[b]{0.2\textwidth}   
        \centering 
        \includegraphics[height=4cm]{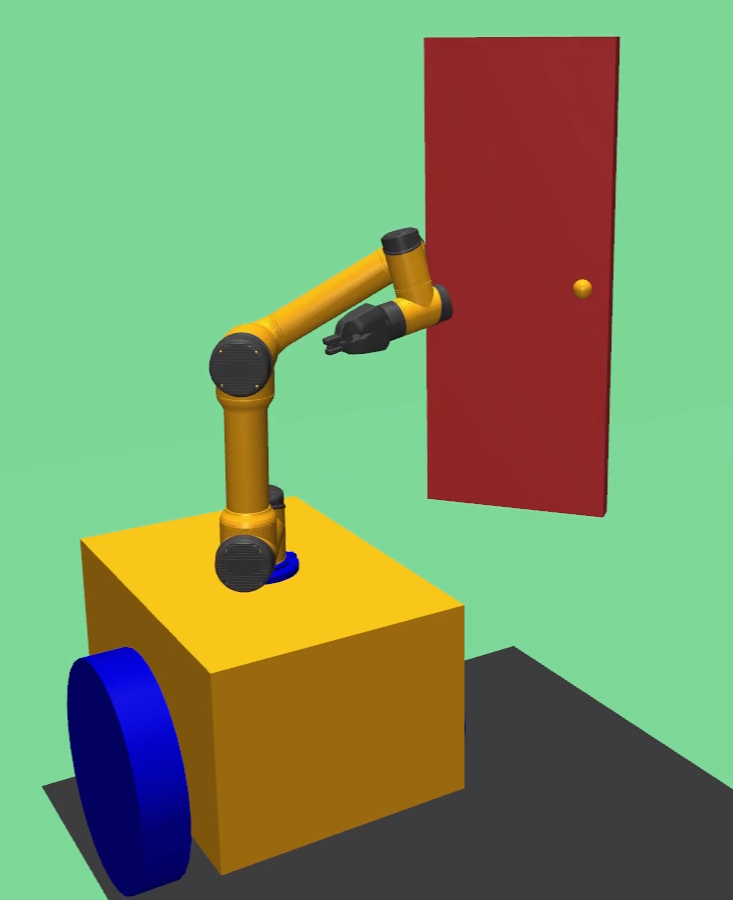}
    \end{subfigure}
	\caption{Robot push open the door.}
    \label{fig:RobotPushDoor}
    \vspace{-10pt}
\end{figure*}

\begin{figure*}[h]
    \begin{subfigure}[b]{0.1\textwidth}
        \includegraphics[width=1.5cm, height=1.5cm]{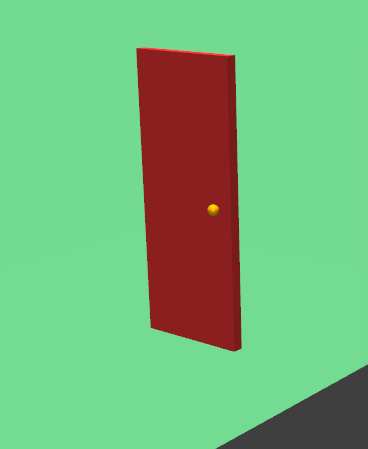}
    \end{subfigure}
    \hspace{\fill}
    \begin{subfigure}[b]{0.1\textwidth}  
        \includegraphics[width=1.5cm, height=1.5cm]{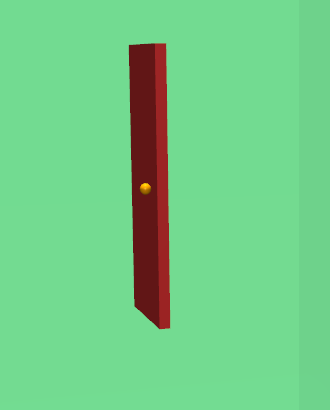}
    \end{subfigure}
    \hspace{\fill}
    \begin{subfigure}[b]{0.1\textwidth}   
        \includegraphics[width=1.5cm, height=1.5cm]{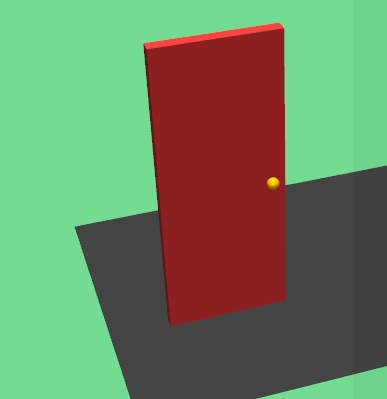} 
    \end{subfigure}
    \hspace{\fill}
    \begin{subfigure}[b]{0.1\textwidth}   
        \includegraphics[width=1.5cm, height=1.5cm]{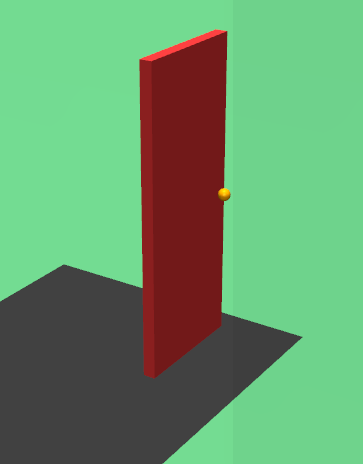}
    \end{subfigure}
   \hspace{\fill}
    \begin{subfigure}[b]{0.1\textwidth}
        \includegraphics[width=1.5cm, height=1.5cm]{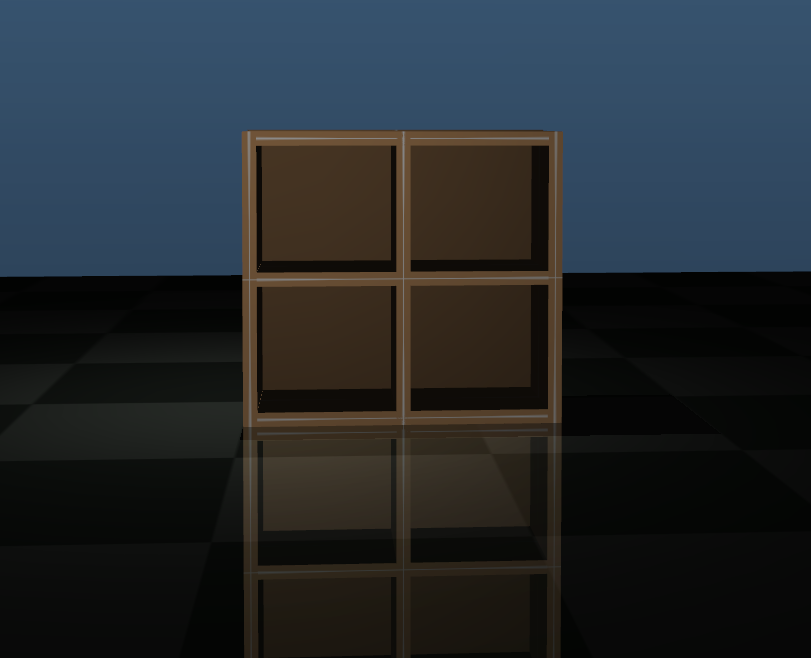}
    \end{subfigure}
    \hspace{\fill}
    \begin{subfigure}[b]{0.1\textwidth}  
        \includegraphics[width=1.5cm, height=1.5cm]{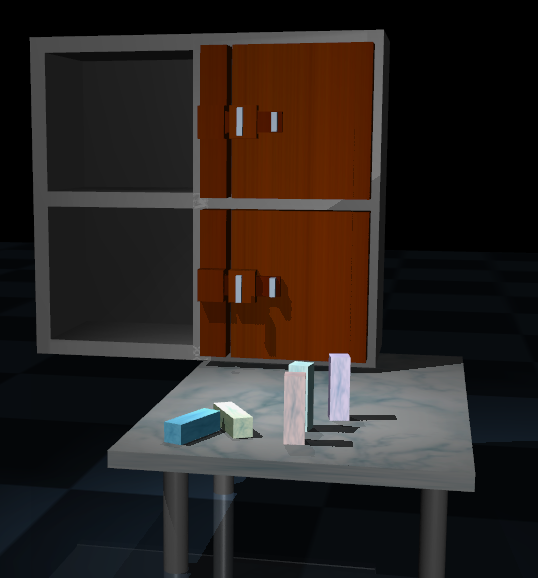}
    \end{subfigure}
    \hspace{\fill}
    \begin{subfigure}[b]{0.1\textwidth}   
        \includegraphics[width=1.5cm, height=1.5cm]{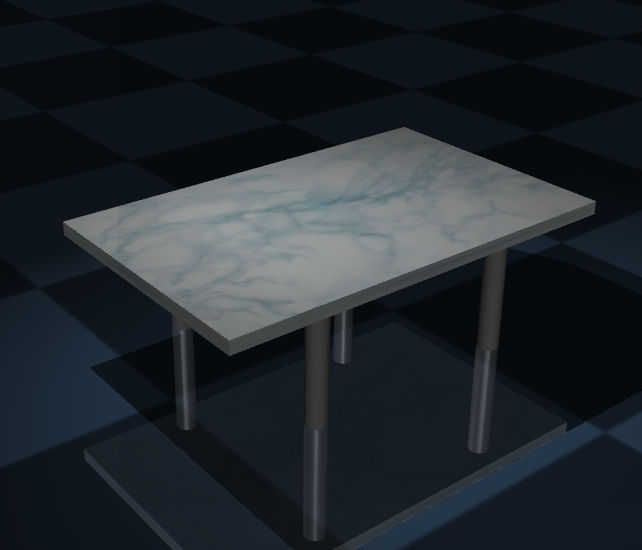} 
    \end{subfigure}
    \hspace{\fill}
    \begin{subfigure}[b]{0.1\textwidth}   
        \includegraphics[width=1.5cm, height=1.5cm]{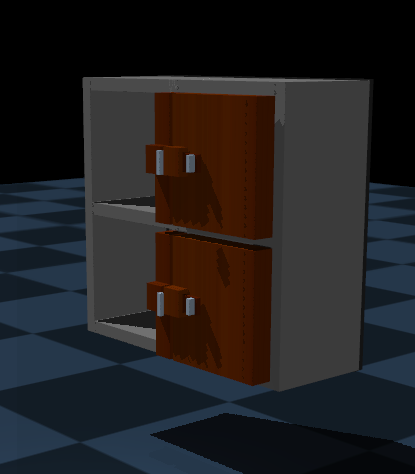}
    \end{subfigure}    
    
    \vspace{0.2cm}
    \begin{subfigure}[b]{0.1\textwidth}
        \includegraphics[width=1.5cm, height=1.5cm]{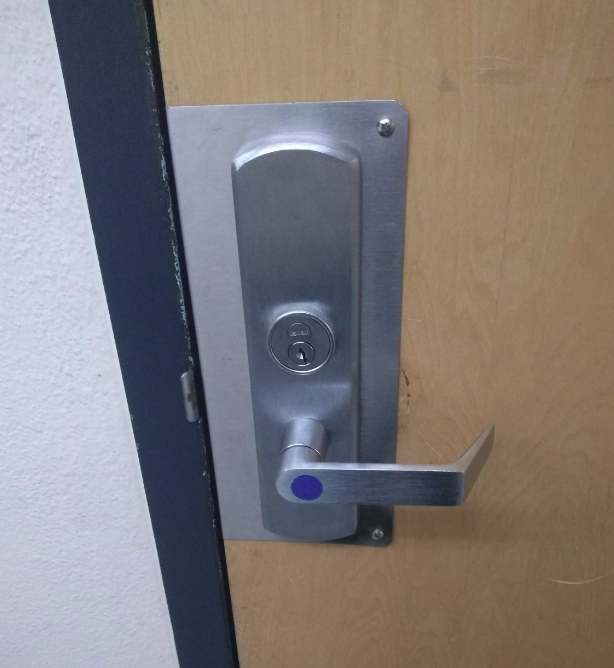}
    \end{subfigure}
    \hspace{\fill}
    \begin{subfigure}[b]{0.1\textwidth}  
        \includegraphics[width=1.5cm, height=1.5cm]{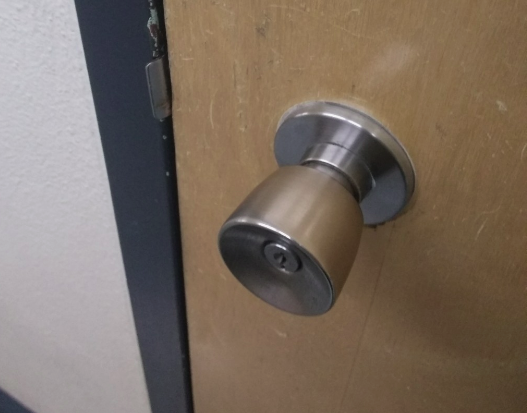}
    \end{subfigure}
    \hspace{\fill}
    \begin{subfigure}[b]{0.1\textwidth}   
        \includegraphics[width=1.5cm, height=1.5cm]{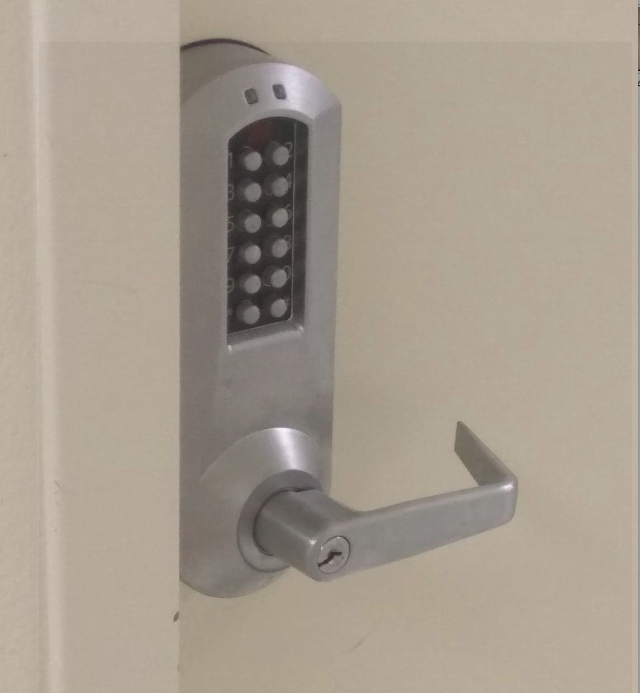} 
    \end{subfigure}
    \hspace{\fill}
    \begin{subfigure}[b]{0.1\textwidth}   
        \includegraphics[width=1.5cm, height=1.5cm]{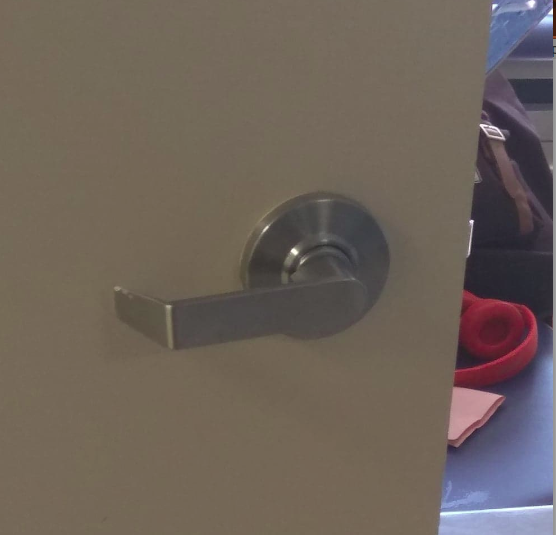}
    \end{subfigure}
   \hspace{\fill}
    \begin{subfigure}[b]{0.1\textwidth}
        \includegraphics[width=1.5cm, height=1.5cm]{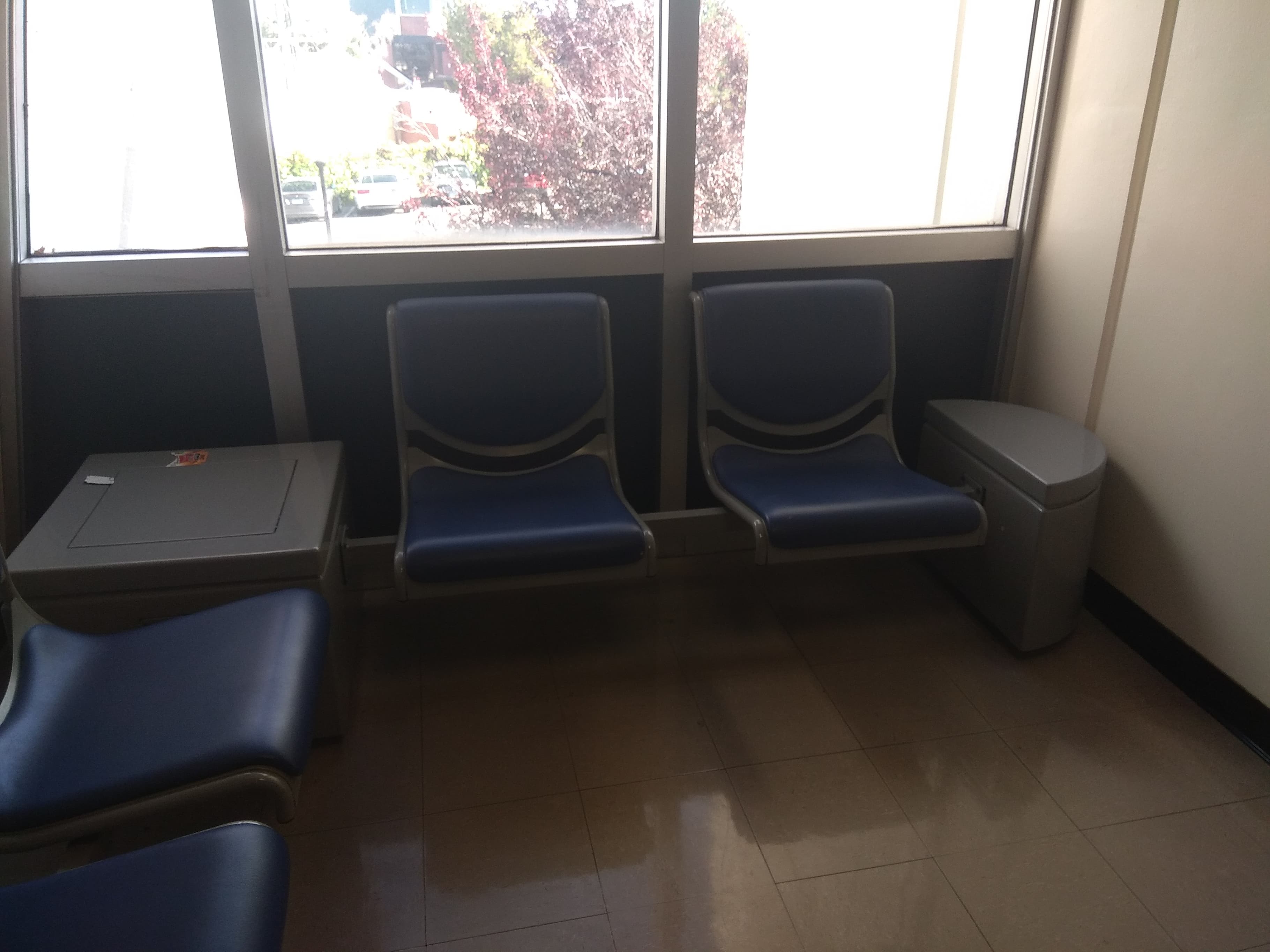}
    \end{subfigure}
    \hspace{\fill}
    \begin{subfigure}[b]{0.1\textwidth}  
        \includegraphics[width=1.5cm, height=1.5cm]{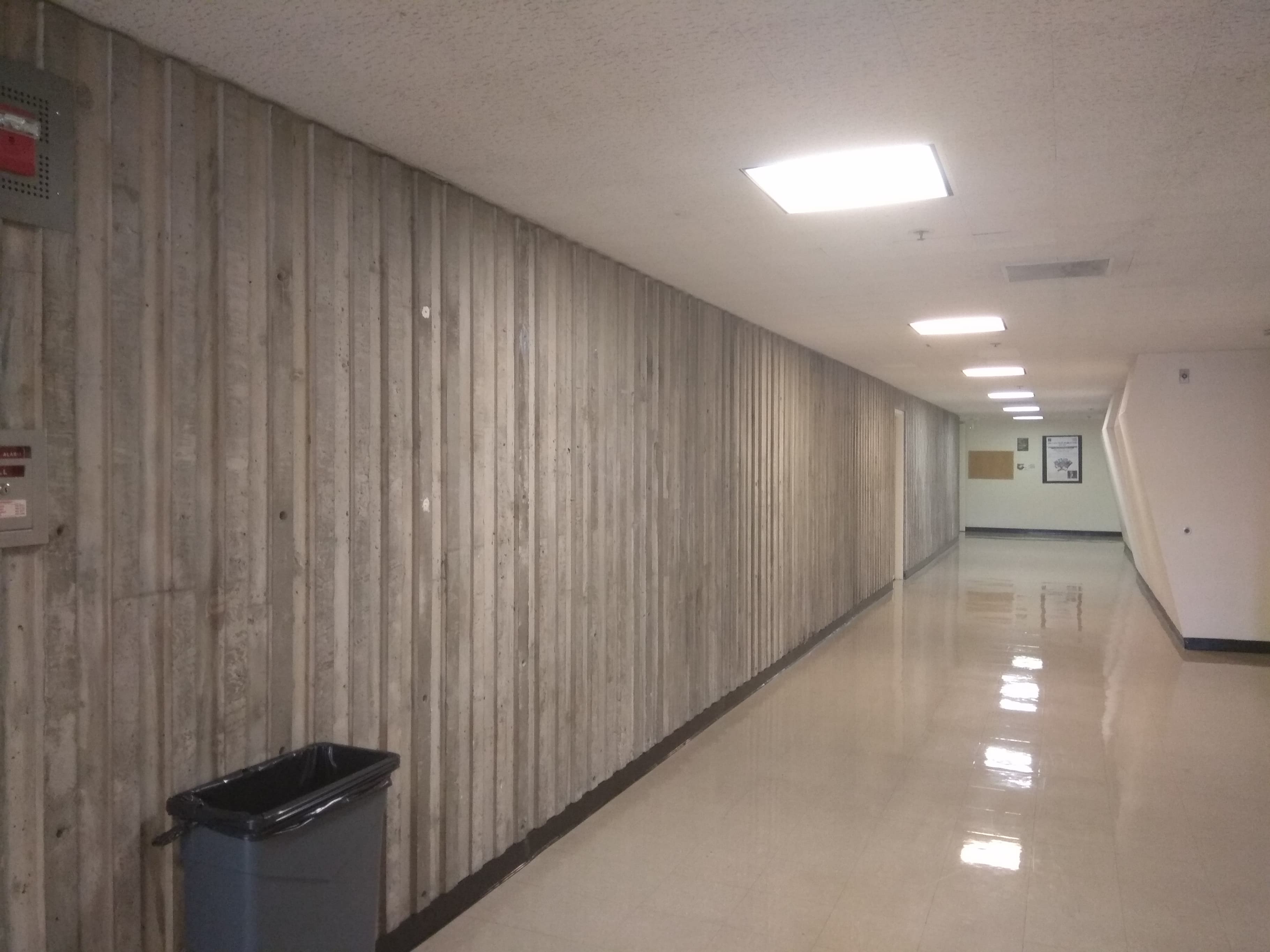}
    \end{subfigure}
    \hspace{\fill}
    \begin{subfigure}[b]{0.1\textwidth}   
        \includegraphics[width=1.5cm, height=1.5cm]{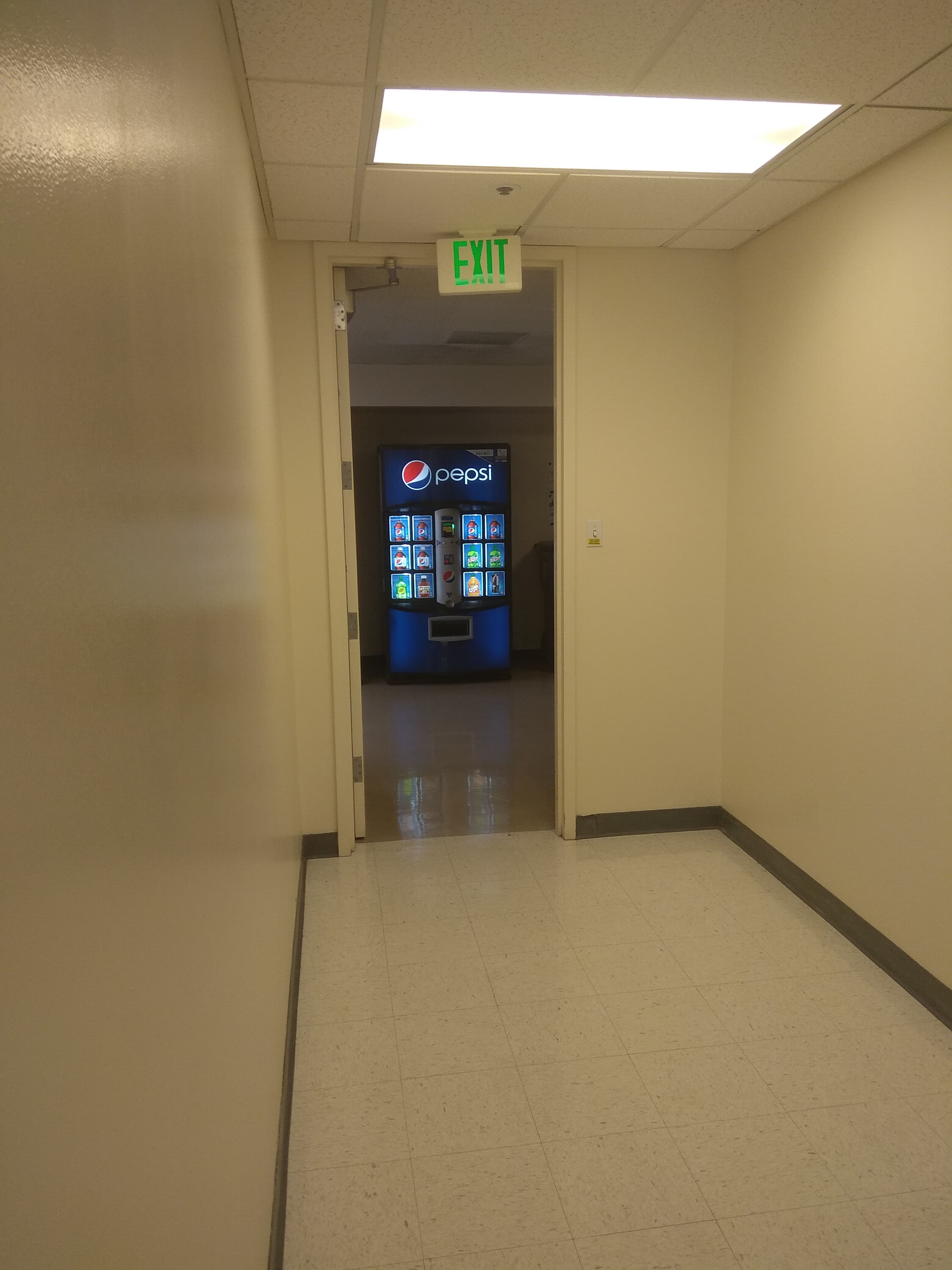} 
    \end{subfigure}
    \hspace{\fill}
    \begin{subfigure}[b]{0.1\textwidth}   
        \includegraphics[width=1.5cm, height=1.5cm]{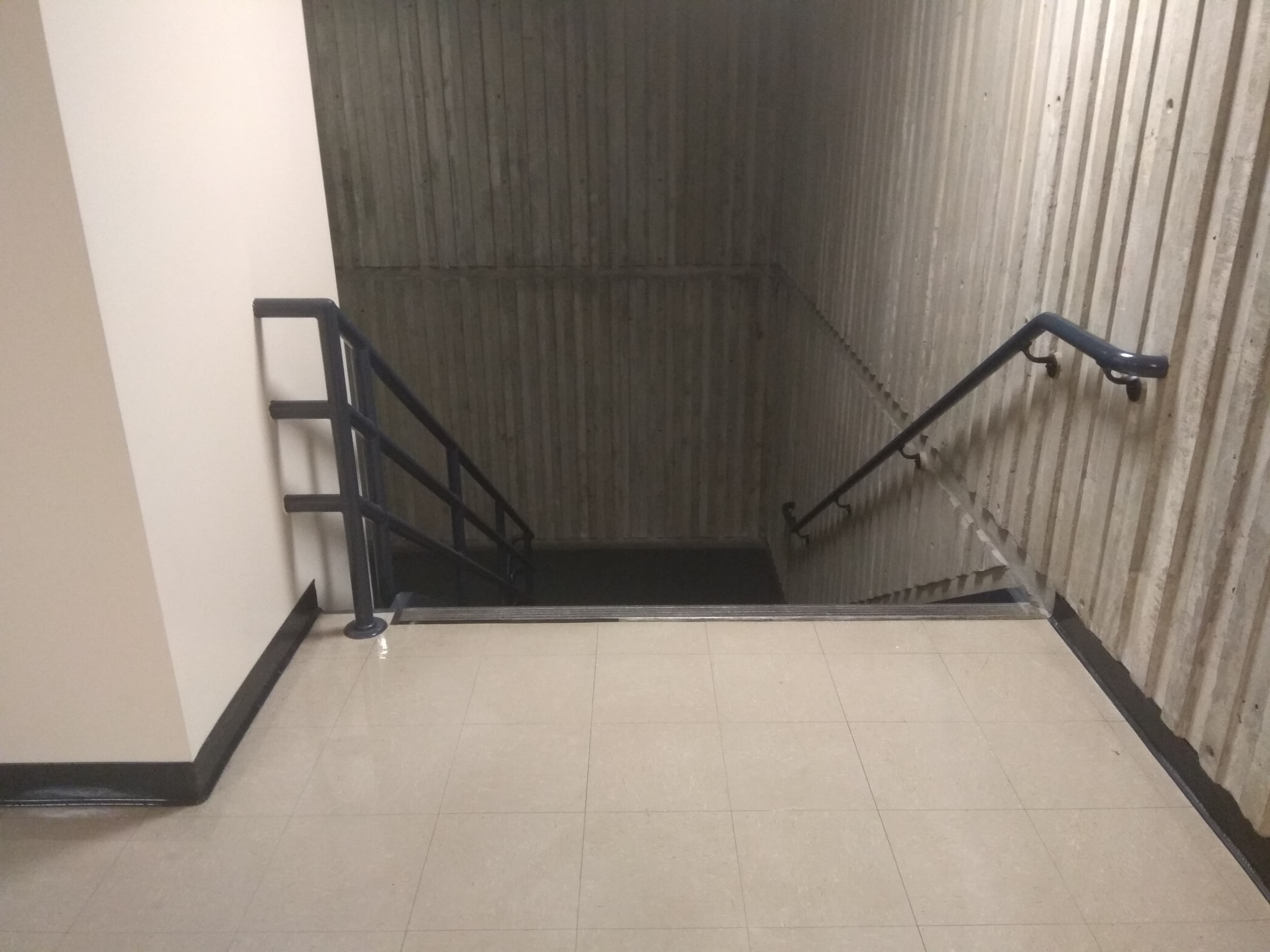}
    \end{subfigure}  
    \caption{Sample training images for door detection training (first row: simulation, second row: real.)}
    \label{fig:CNNImages}
   \vspace{-10pt}
\end{figure*}

We used Gym \cite{Gym} (Robotics Fetch robot environment) and MuJoCo physics simulator \cite{todorov2012mujoco} to perform our experiment. We based on DDPG + HER implementation by OpenAI Baselines \cite{baselines} and modified the \textit{FetchReach-v1} environment to simulate the door opening task. We also verified the improved performance using other three robotics tasks in Gym including: \textit{FetchPush-v1}, \textit{FetchSlide-v1} and \textit{FetchPick\&Place-v1} as shown in Figure \ref{fig:FetchEnvApp}. We created the model for AUBO i5 robot, which we have in our Advanced Robotics and Automation (ARA) lab as shown in Figure \ref{fig:aubo}. It is 6-DoF robotic arm, and it can move using the 4-wheeled mobile base but we keep it static in our experiment. Its joints are controlled by velocities and positions commands. Since the actual robot has built-in gravity compensation, we assume no gravity in the simulation. The complete simulation environment can be seen in Figure \ref{fig:RobotPushDoor}. At each training episode, the location of the door handle (the door also moves along) will be uniformly sampled inside a 30cm x 30cm x 30cm cube with the center fixed in a location. The robot is trained to push open the door using DDPG + HER assisted by ER-CNN.

\subsection{Training Data Generation}
\subsubsection{Experience Ranking CNN}
We use the steps listed in Algorithm \ref{Alg:ER-CNN} to generate training images and then classify them to have labeled data for training. The two inputs are the distance from the robot's end-effector to the door handle and the hinge's angle (to decide whether the door is open or not). The first \textit{while loop} will generate images from training log files and name those images with distance information. The second \textit{while loop} loops over these images, and based on their names (associated distance information), classify images into 5 different groups. The final data set contains in total 1630 images classified in group 0-4. Group 0 images indicate that the manipulator has successfully opened the door at the end of those episodes. Images belonging to Group 4 means that in the episode associated with it, the manipulator did not even touch the door, and the distance from its end-effector to the door handle was the largest. Images in group 1-3 are episodes being in between.

\begin{algorithm}
\caption{Generating labeled images for ER-CNN}\label{euclid}
\begin{algorithmic}[1]
\While{Replay Training Logs} \Comment{Generate images}
\If{hinge angle $>$ 0.05 rad}
\State File name = distance + 'O' + '.png'
\Else
\State File name = distance + '.png'
\EndIf
\EndWhile
\end{algorithmic}
\begin{algorithmic}[1]
\While{Not Last Image} \Comment{Classify generated images}
\State Extract the name
\If{Name contains 'O'}
\State Group 0
\Else
\State Retrieve distance from the name
\State Classify base on the distance
\EndIf
\EndWhile
\end{algorithmic}
\label{Alg:ER-CNN}
 \end{algorithm}

\subsubsection{Door Handle Detection CNN}
In MuJoCo, a camera is integrated to move with the mobile robot's end-effector, and it captures images during the door handle training episodes creating the data set used for training and validating the detection network.

We also experiment with real images, where the training data are acquired from an original set of 180 images of door handles captured around the campus. After that, the data set is augmented by applying augmentation techniques such as horizontal flipping, random rotation, translation or color distortion. The final data set contains 1500 images of door handle and 1484 images of non-door handle images. Some samples of training images can be seen in Figure \ref{fig:CNNImages}. The structure of the CNN shown in Figure \ref{fig:cnnbase}, and it is used for detecting the door handle in both simulation and real images.

\vspace{-10pt}
\subsection{Training and Accuracy}
Both CNNs were trained in an Alienware Aurora R7 computer with an NVIDIA GeForce GTX 1080. Training parameters are listed in Table \ref{Table:CNNTrainParams}. The accuracy achieved for the ER-CNN was about 80.2\% while the second CNN had the accuracy on the test set are 95.1\% and 94.5\% with simulated and outside images, respectively.
\begin{table}[htb]
\centering
 \begin{tabular}{|c | c |} 
 \hline
 \textit{Name} & \textit{Value}\\ [0.5ex] 
 \hline
 Input image & 256x256  \\  
 \hline
 Batch size & 16  \\ 
 \hline
 Sample per epoch & 1000\\
 \hline
 Epochs & 500\\
 \hline
 Learning Rate & 0.001\\
 \hline
 Loss & categorical crossentropy \\
 \hline
 Optimizer &  SGD \\
 \hline 
\end{tabular}
\caption{CNN training parameters.} 
\label{Table:CNNTrainParams}
\vspace{-10pt}
\end{table}

\subsection{Improvement with Experience Ranking}

\begin{figure*}[htb]
    \centering
    \begin{subfigure}[b]{0.45\textwidth}
        \centering
        \includegraphics[width=6.7cm]{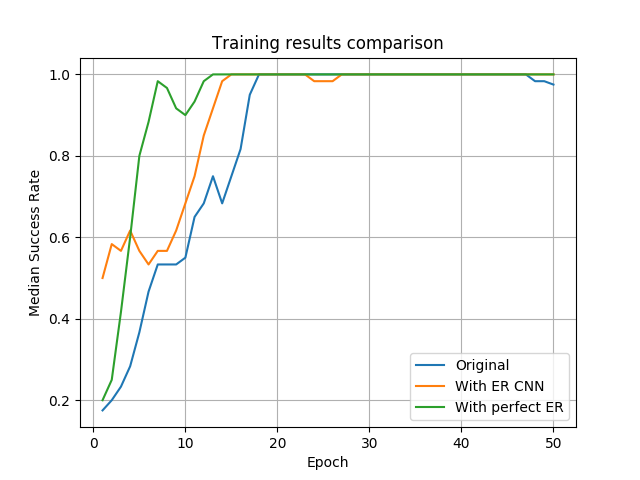}
        \caption{Door Pushing Task.}
    \end{subfigure}
    \begin{subfigure}[b]{0.45\textwidth}  
        \centering 
        \includegraphics[width=6.7cm]{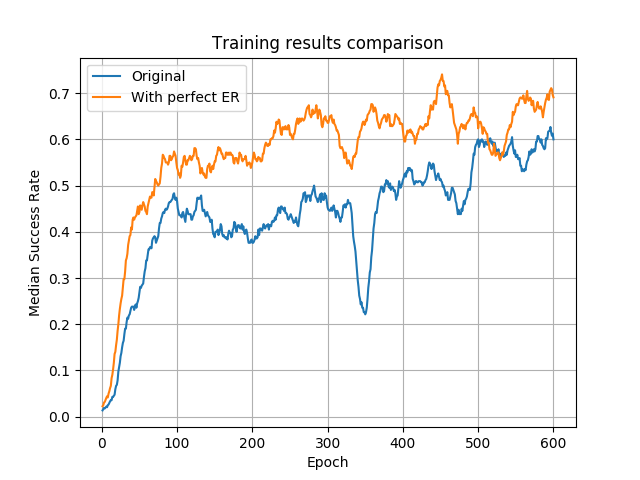}
        \caption{Fetch Slide v1 Task.}
    \end{subfigure}
    \begin{subfigure}[b]{0.45\textwidth}   
        \centering 
        \includegraphics[width=6.7cm]{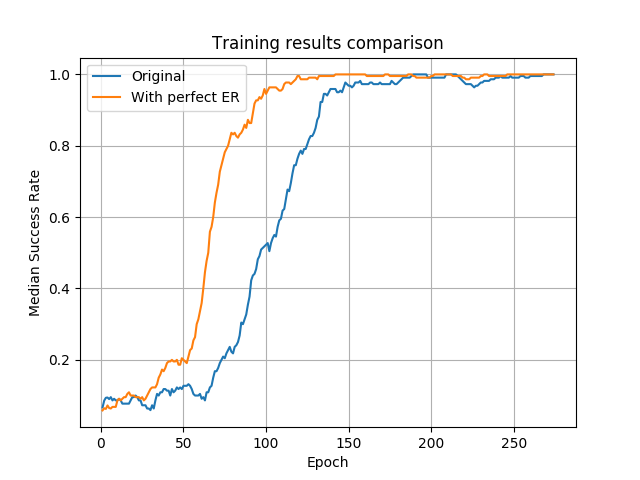} 
        \caption{Fetch Push v1 Task.}
    \end{subfigure}
    \begin{subfigure}[b]{0.45\textwidth}   
        \centering 
        \includegraphics[width=6.7cm]{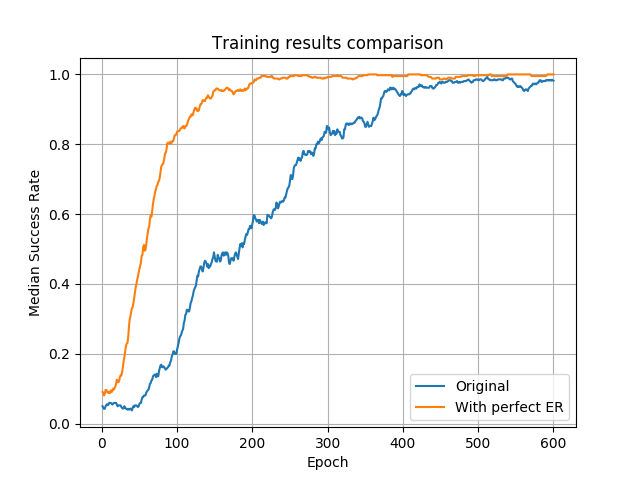}
        \caption{Fetch Pick and Place v1 Task.}
    \end{subfigure}
	\caption{Learning performance comparison in four simulated tasks.}
    \label{fig:FetchEnv}
     \vspace{-10pt}
\end{figure*}
\subsubsection{Door Push training comparison}
In this environment, we first try to measure the improvement when the classifier is 100\% correct to rank the transition. We used entirely the two inputs: the distance from the robot's end-effector and the door handle, and the hinge's angle to correctly classify each episode before deciding to store them. We threw out any experience whose ranks are more than 4, and allow the remaining to enter the replay buffer. After that, we compared the training result with the performance of the original DDPG + HER as shown in Figure \ref{fig:FetchEnv}a. From the figure, we can see that with ER the performance (median success rate) increased more quickly than in case with original DDPG + HER. More interestingly after convergence, the success rate stayed at 100\% while the original DDPG + HER still struggles to maintain the same accuracy. This phenomenon might indicate that supervised learning actually helps stabilize the learning in particular when the learning converges. We also implement the trained ER-CNN to replace the perfect classifier and compare the results. As expected, the performance decreases but it still converges faster and more robust than using original DDPG + HER.
\subsubsection{Other tasks} Because door pushing is quite a simple task for DDPG + HER, we continue exploring the results in other more complex existing tasks in the Gym Robotics environment including: \textit{FetchPick{\&}Place-v1}, \textit{FetchPush-v1} and \textit{FetchSlide-v1}. In these tasks, we only perform the comparison between the perfect classifier (DDPG + HER + ER) and the original,  without ER (DDPG + HER). From Figure \ref{fig:FetchEnv}b,c,d we can clearly see the improved learning performance gained from using ER, the learning converges faster and more robust after convergence than the original DDPG + HER. Our method took only half the number of episodes needed for the original DDPG + HER for convergence in \textit{FetchPush-v1} and a third in \textit{FetchPick\&Place-v1} environment.

\section{CONCLUSION AND FUTURE WORK}
\label{Content:Conclusion}
We have proposed a new method to improve performance of existing DDPG + HER combination by introducing experience ranking using CNN for each episode to decide whether it should be stored and replayed. We also validated the proposed methodology with four different robotics simulated tasks. By just throwing out the lowest rank experience, the learning performance outperformed the original DDPG + HER in all tasks. Therefore, experience ranking using an external evaluation element can speed up learning, and it is easily extended to other more complex RL problems as well. We believe that by embedding task-relevant knowledge within robots (in our cases in forms of pre-trained CNNs), they will learn faster just like the way human learn a new task. Finally, perhaps to learn faster, we should focus more on the quality of training data rather than the quantity.


%


\bibliography{bibliographies/reference.bib}
\bibliographystyle{IEEEtran}
\end{document}